\documentclass[11pt,a4paper]{article}
\usepackage[hyperref]{eacl2021}
\usepackage{times}
\usepackage{latexsym}
\usepackage{booktabs}
\usepackage{xargs}
\usepackage{tikz-dependency}

\usepackage{microtype}

\usepackage{tikz}
\usepackage{subcaption}
\usepackage{booktabs}
\usepackage{paralist}
\usepackage{xspace}
\usepackage{arydshln}

\usepackage{gb4e}
\noautomath     % otherwise gb4e causes compiler errors
\definecolor{BrickRed}{RGB}{105, 21, 32}
\definecolor{ForestGreen}{RGB}{224, 222, 25}
\definecolor{myBlue}{RGB}{69, 162, 217}

\newcommand{\tr}{\texttt{TR}\xspace}
\newcommand{\DE}{\texttt{DE}\xspace}
\newcommand{\other}{\texttt{Other}\xspace}
\renewcommand{\ne}{\texttt{NE}\xspace}
\newcommand{\lang}{\texttt{Lang3}\xspace}
\newcommand{\mixed}{\texttt{Mixed}\xspace}
\newcommand{\ambig}{\texttt{Ambig}\xspace}

\newcommand{\trde}{Tr-De\xspace}
\newcommand{\iden}{Id-En\xspace}

\newcommand{\ind}{\texttt{ID}\xspace}
\newcommand{\en}{\texttt{EN}\xspace}
\newcommand{\un}{\texttt{UN}\xspace}

\newcommand{\raw}{\texttt{Raw}\xspace}
\newcommand{\tokanon}{\texttt{Tok+Anon}\xspace}
\newcommand{\norm}{\texttt{Norm}\xspace}
\newcommand{\segcs}{\texttt{Seg+CS}\xspace}

\newcommand{\ltag}[1]{\textsc{#1}}

\aclfinalcopy 
\def\aclpaperid{912} 

\title{Lexical Normalization for Code-switched Data and its Effect on POS
Tagging}

\author{Rob van der Goot \\
  IT University of Copenhagen \\
  {\tt robv@itu.dk} \\\And
  {\"O}zlem \c{C}etino{\u{g}}lu  \\
  IMS, University of Stuttgart \\
  {\tt ozlem@ims.uni-stuttgart.de} \\}

\date{}

\begin{document}
\maketitle
\begin{abstract}
Lexical normalization, the translation of non-canonical data to standard
language, has shown to improve the performance of many natural language
processing tasks on social media.  Yet, using multiple languages in one
utterance, also called code-switching (CS), is frequently overlooked by these
normalization systems, despite its common use in social media.  In this paper,
we propose three normalization models specifically designed to handle
code-switched data which we evaluate for two language pairs: Indonesian-English
(\iden) and Turkish-German (\trde). For the latter, we  introduce novel
normalization layers and their corresponding language ID and POS tags for the
dataset, and evaluate the downstream effect of normalization on POS tagging.
Results show that our CS-tailored normalization models outperform \iden state
of the art  and \trde monolingual models, and lead to 5.4\% relative
performance increase for POS tagging as compared to unnormalized input.
\footnote{Source code is available at:
\url{https://bitbucket.org/robvanderg/csmonoise}.  The Turkish-German data is
available at: \url{https://github.com/ozlemcek/TrDeNormData}}
\end{abstract}

\section{Introduction}
Social media provide an invaluable source of information for natural language
processing (NLP) systems. Its informative and spontaneous nature leads to many
interesting phenomena, like non-standard words, spelling errors and
abbreviations. One particularly challenging and interesting phenomenon is the
use of multiple languages within the same utterance, which is also called
code-switching (CS)~\cite{gumperz1982discourse,myers1995social,toribio2012}.

Because most NLP models are designed to process canonical and monolingual data,
their performance drops enormously when having to process social media
data~\cite{eisenstein-2013-bad}. One solution to this problem is lexical
normalization: the translation of non-standard (e.g. social media) text to its
canonical form~\cite{han-baldwin-2011-lexical}. Previous work has shown that by
standardizing the data, we can improve the robustness of NLP
systems~\cite{derczynski-etal-2013-twitter,zhang-etal-2013-adaptive,van-der-goot-van-noord-2017-parser}.
Nevertheless these systems overlook code-switching.  (\ref{ex:iden.norm}) shows
a code-switched tweet (upper) and its normalization annotation (lower), taken
from an Indonesian-English CS corpus \cite{barik-etal-2019-normalization}
(Indonesian in bold). This example demonstrates that CS complicates
normalization, because it can be unclear in which language to normalize (e.g.,
\textit{ak} is normalized to \textit{aku} `I' in Indonesian. English-only
normalization systems would probably normalize it to \textit{ok}).

\begin{exe}
%\begin{small}
\ex\label{ex:iden.norm} 
 \gll  \textbf{ak}  .  luv  u  :(  till  die \\
       \textbf{aku}  .  love  you  :(  till  die\\
 \trans `\textit{I love you till (I) die}'
%\end{small} 
\end{exe}

Recently, there has been an increasing interest in the automatic processing of
CS data, however, there has not been much work on its lexical normalization.
To the best of our knowledge, only \newcite{adouane-etal-2019-normalising}
focus entirely on lexical normalization for CS data in their work.  For other
works, normalization is a preprocessing step for downstream tasks: chunking
\cite{sharma:2016},
parsing~\cite{bhat-etal-2017-joining,bhat-etal-2018-universal}, or machine
translation ~\cite{barik-etal-2019-normalization}. These CS normalizers are
either rule-based and language-specific \cite{barik-etal-2019-normalization} or
combine (Hindi) back-transliteration and normalization
\cite{sharma:2016,bhat-etal-2017-joining,bhat-etal-2018-universal} thus, they
are not directly applicable to other lexical normalization datasets. In this
work,  
\begin{itemize}
    \item We are the first to present open-source normalization models
specialized for CS lexical normalization without any language-specific
components. 
    \item We provide a novel lexical normalization dataset by annotating a
Turkish-German Twitter corpus \cite{cetinoglu-2016-turkish}. We also align
existing annotation layers -- language IDs (LID) and part-of-speech (POS) tags
-- to normalization annotations.
    \item We evaluate three CS normalization models on two language pairs
(Turkish-German (\trde), Indonesian-English (\iden)). For both datasets, CS
models reach performance in a similar range as monolingual models reach on
monolingual datasets. 
    \item Our CS-tailored normalization models outperform \iden state of the
art  and set the state of the art for the \trde dataset. 
    \item We show that our proposed normalization models improve the
performance of POS taggers. For a broad perspective, we employ a variety of
taggers (CRF, BiLSTM, BERT).
\end{itemize}

\section{Related Work}
\paragraph{Lexical normalization}
Traditionally, social media normalization approaches can be broadly divided
into two types.  The first stream of work uses techniques borrowed from machine
translation~\cite{aw-etal-2006-phrase,pennell-liu-2011-character,ljubevsic2016normalising}.
The second stream is based on a classic spelling correction framework
(noisy-channel models)~\cite{han2014improving}.  Here, they often apply three
steps, detecting which words need to be replaced, generating candidates, and
ranking these candidates. Later, it became evident that a two-step approach is
sufficient~\cite{jin:2015:WNUT,van-der-goot-2019-monoise}, and the detection
step was alleviated by considering the original word as a normalization
candidate.

The current state-of-the-art model for most languages is
MoNoise~\cite{van-der-goot-2019-monoise}, which is based on this two-step
approach. A variety of modules are used for the generation of candidates. For
the ranking, MoNoise complements features from the generation step with
additional features, which are all combined in a random forest classifier that
predicts the probability that a candidate is a `correct' candidate. MoNoise is
described in more detail in Section~\ref{sec:monoise}.  More recently,
sequence-to-sequence models~\cite{lourentzou2019adapting} and contextual
embeddings~\cite{muller-etal-2019-enhancing} have been used for the lexical
normalization task. These approaches have been shown to reach performances
close to MoNoise on English benchmarks.

Like most NLP tasks, most research on normalization has been done on English
datasets~\cite{han-baldwin-2011-lexical,baldwin-etal-2015-shared}. However,
there has been some efforts on other languages, where usually only one language
is considered, we refer to~\newcite{sharf2017lexical}
and~\newcite{van-der-goot-2019-monoise} for an overview of available resources.

\paragraph{Processing of code-switched social media data}
Early work on normalizing CS data focused on Hindi-English, as part of
pipelines to achieve downstream tasks
\cite{sharma:2016,bhat-etal-2017-joining,bhat-etal-2018-universal}.  As Hindi
is Romanized in datasets and additional Hindi resources are in the Devanagari
script, they include back-transliteration into the normalization step, thus
defining the task beyond the scope of this paper. Nevertheless, all systems
report a positive impact of normalization on their final task.

More recently, ~\newcite{barik-etal-2019-normalization} experiment on
normalization for Indonesian-English.  They use a rule-based approach
supplemented by clusters derived from word embeddings, and show that
normalization can be used to improve machine translation.
\newcite{adouane-etal-2019-normalising} instead propose to use
sequence-to-sequence models for normalizing Algerian Arabic data mixed with
Modern Standard Arabic, French, Berber, and English.  They show that their edit
distance-based token-level aligner helps improve normalization.

\renewcommand{\thefigure}{2}

{
\begin{figure*}
\centering
\begin{tabular}{p{1cm}llclcl}
 & \multicolumn{2}{c}{\segcs} & & \multicolumn{1}{c}\norm & & \multicolumn{1}{c}\tokanon \\
 \cline{2-3} \cline{5-5} \cline{7-7}
Tokens & Semester{\S}da & -y{\i}m & & Semesterday{\i}m & & semesterdayim \\
LID & \mixed & \tr & \hspace{0.3cm}$\Rightarrow$\hspace{0.3cm} &  \mixed & \hspace{0.3cm}$\Rightarrow$\hspace{0.3cm} & \mixed \\
POS & \ltag{noun} & \ltag{verb} & & \ltag{verb} & &  \ltag{verb}\\
\end{tabular}
\caption{\label{fig:map}Mapping LID and POS tags from \segcs to \norm to \tokanon for the mixed word \textit{Semesterday{\i}m} `I am in semester'.}
\end{figure*}
}

\renewcommand{\thefigure}{1}

{ \setlength\tabcolsep{3pt}
\begin{figure}
 \resizebox{\linewidth}{!}{

\centering
\begin{tabular}{ll|l|l|l|l|l}
\raw: & \multicolumn{5}{l}{@Erkan1903 nerdee 3 \textbf{semester}dayim dha.}\\
\tokanon: & @username & nerdee & 3 & \textbf{semester}dayim & dha & .\\
\norm & @username & Nerde & 3. & \textbf{Semester}day{\i}m & daha & .\\
\segcs: & \multicolumn{5}{l}{@username Nerde 3. \textbf{Semester}{\S}da -y{\i}m daha .}\\
\end{tabular}
}
\caption{\label{fig:prep} Different annotation layers of a tweet from the \trde
corpus, meaning `\textit{No way, I am still in the 3rd semester}'. The German
part is in bold. \raw: downloaded tweet; \tokanon: after tokenization and
anonymization; \norm: after normalization; \segcs: after segmentation (e.g. the
Turkish copular \textit{-yim}), and CS boundaries ({\S}) in \mixed tokens. The
token alignment and normalization tasks are carried out on the \tokanon and
\norm pairs.}
\end{figure}
}

\renewcommand{\thefigure}{3}

When annotating the \trde dataset for normalization, we also adapted its POS
tags (see Section \ref{sec:trdedata}).  This gives us the opportunity to apply
POS tagging as extrinsic evaluation.  Besides research on Hindi-English that
combines normalization and back-transliteration, most work either use
normalization to improve tagging performance of monolingual social media
data~\cite{derczynski-etal-2013-twitter,vandergoot-plank-nissim:2017:WNUT}, or
on POS tagging of CS data without
normalization~\cite{alghamdi-etal-2016-part,soto-hirschberg-2018-joint}. In
this work, we combine these angles.

Because some of our proposed normalization models depend on language labels, we
require a word-level language identification system. There is a wide variety of
approaches used for this task, where early systems mostly used
CRFs~\cite{sequiera2015overview,molina-etal-2016-overview}. More recently,
neural networks based approaches  %for this task have shown superior
performance for this task~\cite{zhang-etal-2018-fast}.  We opt for three
different architectures to observe the effect of the quality of language
identification on normalization (Section~\ref{sec:method:lang}).

\section{Data}
In this section we first describe the design decisions of the novel
Turkish-German dataset, then we compare some basic statistics together with the
existing Indonesian-English dataset~\cite{barik-etal-2019-normalization}.

\subsection{Turkish-German code-switched normalization corpus}
\label{sec:trdedata}
We use the Turkish-German Twitter corpus from~\newcite{cetinoglu-2016-turkish}
in our experiments.  It consists of 17K tokens as 1,029 tweets. The raw tweets
of the corpus have undergone three main steps of alternations after the
collection: tokenization, normalization, and
segmentation.\footnote{Morphosyntactic split of words into subwords, cf.
\cite{cetinoglu-coltekin-2016-part} for details.} In addition, usernames and
URLs are anonymized as \texttt{@username} and  \texttt{[url]} respectively, and
intra-word CS boundaries are marked in \mixed tokens with §. Each alternation
layer is exemplified on a sentence from the corpus in Figure \ref{fig:prep}. 

The \segcs layer is annotated with language IDs and POS tags
\cite{cetinoglu-coltekin-2016-part}. The LID tag set consists of \tr (Turkish),
\DE (German), \lang (third language), \mixed (intra-word CS), \ne (named
entity), \ambig (both Turkish and German and cannot be disambiguated in given
context), \other (punctuation, numbers, URLs, emoticons, symbols).
Additionally, named entities are tagged with their language label next to the
\ne tag, e.g. `Germany' is annotated in the corpus as follows depending on the
language: \textit{Almanya} \ne.\tr, \textit{Deutschland} \ne.\DE,
\textit{Germany} \ne.\lang.  The POS annotation adopts the Universal
Dependencies (UD) tag set~\cite{NivreUDV16}. 

\paragraph{Preprocessing for normalization}
The original version of the corpus has only the \raw and \segcs layers and only
tweet-level alignment between them. As our work focuses only on normalization
we created the intermediate layers \tokanon and \norm that leave out other
tasks. Since MoNoise requires word-aligned annotations, we also provided these
alignments. 

We anonymized and tokenized the raw tweets to achieve the \tokanon layer.  For
tokenization, we use a slightly modified version of
twokenize.py\footnote{\url{github.com/brendano/tweetmotif}}
\cite{o2010tweetmotif}.  To obtain the \norm layer, we merged back segmented
tokens and removed CS boundaries on the \segcs layer. 

After this stage, we aligned \tokanon and \norm on the token level
automatically using Giza++ \cite{och03:asc}.  We parsed the resulting alignment
files to align the actual tokens and corrected them manually. There are 15,715
1:1, 520 1:n, and 147 n:1 alignments.

\paragraph{LID and POS alignment}
The existing LID and POS tags are on the \segcs layer; since we base our
experiments on the \tokanon layer, we need to map the annotations.  This is
done in two steps following the \segcs $\Rightarrow$ \norm $\Rightarrow$
\tokanon order. Due to segmentation merges in the first step, and 1:n and n:1
token alignments in the second step, there are non-trivial LID and POS
alignments. 

Figure \ref{fig:map} demonstrates a segmented word in the first column. The
first segment \textit{Semesterda} `in semester' is \mixed with German
\textit{Semester} and Turkish locative case marker \textit{da}. The second
segment is the Turkish copular \textit{-yim} `I am'. Their POS tags are
\ltag{noun} and \ltag{verb}, respectively. When segmentation is undone in the
second column (\norm), their LID and POS are merged too. If two tokens have the
same LID, the merged token takes the same LID. If they are different, the
resulting token is \mixed, as in the example.

POS tag merging rules can get more complicated, therefore, we used a heuristic
that favors the POS tag of the second token in most cases.\footnote{Turkish is
agglutinative. Segmentation often happens by splitting derivational suffixes
that bear the final POS tag.} When a \ltag{noun} segment is merged with a
\ltag{verb} segment, as in Figure 2 (\segcs $\Rightarrow$ \norm), the merged
token is assigned a \ltag{verb} POS tag.  For the \norm $\Rightarrow$ \tokanon
mapping, the alignment is 1:1, thus LID and POS are directly carried over.

 \begin{table}
 \centering
 \resizebox{\linewidth}{!}{
\begin{tabular}{l | r r r r r r r r r}
\toprule
        & \#words & \%norm & \% split & \%merge &  CMI\\
        \midrule
Id-En & 18,758 & 14.13 & 1.33 & 0.17 &  28.20 \\
Tr-De &  13,217 & 25.97 & 3.01 & 1.04 & 22.44 \\

\bottomrule
\end{tabular}}
\caption{Descriptive normalization and code-switching statistics on the
training split of the datasets. CMI is the code-mixing
index~\cite{das-gamback-2014-identifying}, averaged over all training
sentences. \%norm reflect the percentage of words which is normalized.}
\label{tab:datasets}
\end{table}

\subsection{Dataset characteristics}
\label{sec:data}
Besides the data described in the previous section, we use the
Indonesian-English (\iden) data from~\cite{barik-etal-2019-normalization}.  The
\iden data is only annotated with language IDs and uses three labels: \ind,
\en, \un (Unspecified), whereas the \trde includes 12 labels
(Section~\ref{sec:trdedata}). To simplify the models and improve comparability,
we map the language labels of the \trde dataset  to \tr, \DE and \un. Named
entities are  mapped to their respective language tags, e.g, \ne.\DE to \DE.
\mixed tokens are mapped to \DE as they are German words with Turkish
inflection. \lang, \ambig and \other are mapped to \un.

We divide both datasets into a train and test split (80-20\%), and omit a
development set due to small sizes. Since we want to leave test set out in
analyses, we opt for 10-fold cross-validation on the training split of the data
in experiments.  Statistics of the training splits of the datasets are shown in
Table~\ref{tab:datasets}.  The datasets are relatively small, but a high ratio
of words is normalized, including a high percentage of splits and merges.  The
percentage of in-vocabulary words is especially low in the \trde data, which is
mainly due to the morphological richness of Turkish. The code-mixing index
(CMI)~\cite{das-gamback-2014-identifying} indicates the (average of the) amount
of words not written in the majority language for each sentence. The relatively
high CMI  for both datasets indicates a high frequency of code-switching occurs
in the data.

In both datasets there are a small amount of sentences without normalization (8
and 76 for respectively \iden and \trde), which might be desirable for
evaluation of (over)normalization, as in a real-world setup one also does not
know beforehand whether normalization is necessary. In more than half of the
sentences the number of normalized words is larger than 3.  Furthermore, there
are some sentences (5-10 per dataset) with a very high normalization ratio
($>$70\%), which are all in capitals. 

\subsection{Monolingual Data}
\label{sec:rawData}
Our baseline model (MoNoise) exploits monolingual data from both the source and
the target domain (canonical data) to train word embeddings and estimate n-gram
probabilities. To this end, we utilize Wikipedia dumps from 01-01-2020 and
random tweets collected throughout 2012 and 2018 from the Twitter API, filtered
by the FastText language classifier~\cite{joulin2017bag}. We tokenized this
data based on whitespaces, and removed all duplicate sentences/tweets. The
sizes of the collected raw datasets are shown in Table~\ref{tab:datasets2}.

\begin{table}
\centering
 \resizebox{\linewidth}{!}{

\begin{tabular}{l |r r r r}
    \toprule
    Source & Indonesian & English & Turkish & German \\
    \midrule
    Wikipedia  & 75 & 2,162 & 55 & 776 \\
    Twitter & 510 & 5,018 & 203 & 89 \\
    \bottomrule
\end{tabular}}
\caption{Size of raw data (in million words) from both data sources.}
\label{tab:datasets2}
\end{table}

\section{Models}
\begin{figure*}
\centering
\vspace{.2cm}
 \resizebox{.9\textwidth}{!}{
    \begin{subfigure}{.2\textwidth}
        \begin{tikzpicture}
    \path[use as bounding box] (-1,-.5) rectangle (1,4.5);

    %\draw [myBlue, line width=.05cm] (0,4) ellipse (.65cm and .35cm);
    %\draw [myBlue, line width=.05cm] (0,4) ellipse (.65cm and .35cm);
    \node (classifier1) [rectangle, line width=.05cm, draw, myBlue, minimum width=.5cm, minimum height=.5cm, rotate=45] at (0,4) {};

    \node (classifier) at (0,3.75) {};
    \node (feats1) [rectangle, line width=.05cm, draw, ForestGreen, minimum width=.75cm, minimum height=.75cm] at (-.5,2) {};
    \node (feats2) [rectangle, line width=.05cm, draw, myBlue, minimum width=.75cm, minimum height=.75cm] at (.5,2) {};

    \node (input) [] at (0,0) {Input};
    
    %\draw[->] input -| feats1;
    \draw [->] (input) -- (feats1);
    \draw [->] (input) -- (feats2);
    \draw [->] (feats1) -- (classifier);
    \draw [->] (feats2) -- (classifier);
    \draw [->] ([yshift=.5cm]classifier.north) -- ([yshift=.75cm]classifier.north);

\end{tikzpicture}
        \caption{\texttt{Monolingual} MoNoise.}
        \label{fig:monoise}
    \end{subfigure}
    \hspace{.3cm}
    \begin{subfigure}{.2\textwidth}
        \begin{tikzpicture}
    \path[use as bounding box] (-1.5,-.5) rectangle (1.5,4.5);
        
    %\draw [myBlue, line width=.05cm] (-.75,4) ellipse (.65cm and .35cm);
    \node (classifier11) [rectangle, line width=.05cm, draw, ForestGreen, minimum width=.5cm, minimum height=.5cm, rotate=45] at (-.75,4) {};
    \node (classifier1) at (-.75,3.75) {};
    \node (classifier21) [rectangle, line width=.05cm, draw, BrickRed, minimum width=.5cm, minimum height=.5cm, rotate=45] at (.75,4) {};
    %\draw [myBlue, line width=.05cm] (.75,4) ellipse (.65cm and .35cm);
    \node (classifier2) at (.75,3.75) {};

    \node (feats1) [rectangle, line width=.05cm, draw, ForestGreen, minimum width=.75cm, minimum height=.75cm] at (-1,2.5) {};
    \node (feats2) [rectangle, line width=.05cm, draw, myBlue, minimum width=.75cm, minimum height=.75cm] at (0,2.5) {};
    \node (feats3) [rectangle, line width=.05cm, draw, BrickRed, minimum width=.75cm, minimum height=.75cm] at (1,2.5) {};

    \draw [line width=.05cm] (0,1) ellipse (.375cm and .375cm);
    \node (lang)[rectangle, minimum width=.7cm, minimum height=.7cm] at (0,1) {};

    \node (input) [] at (0,0) {Input};
    
    %\draw[->] input -| feats1;
    \draw [->] (input) -- (lang);
    \draw [->, dashed] (lang) -- (feats1);
    \draw [->] (input) .. controls ++(-0.65, 1) and ++(-.65, -1)  ..  (feats2);

%    \draw [->] (lang) -- (feats2);
    \draw [->, dashed] (lang) -- (feats3);
    \draw [->] (feats1) -- (classifier1);
    \draw [->] (feats3) -- (classifier2);
    \draw [->] (feats2) -- (classifier1);
    \draw [->] (feats2) -- (classifier2);
    \draw [->] ([yshift=.5cm]classifier1.north) -- ([yshift=.75cm, xshift=.5cm]classifier1.north);
    \draw [->] ([yshift=.5cm]classifier2.north) -- ([yshift=.75cm, xshift=-.5cm]classifier2.north);

\end{tikzpicture}
        \caption{\texttt{Fragment} based $\;$ normalization.}
        \label{fig:fragment}
    \end{subfigure}
    \hspace{.3cm}
    \begin{subfigure}{.2\textwidth}
        \begin{tikzpicture}
    \path[use as bounding box] (-1.5,-.5) rectangle (1.5,4.5);

    %\draw [myBlue, line width=.05cm] (0,4) ellipse (.65cm and .35cm);
    \node (classifier21) [rectangle, line width=.05cm, draw, myBlue, minimum width=.5cm, minimum height=.5cm, rotate=45] at (0,4) {};

    \node (classifier) at (0,3.75) {};
    \node (feats1) [rectangle, line width=.05cm, draw, ForestGreen, minimum width=.75cm, minimum height=.75cm] at (-1,2) {};
    \node (feats2) [rectangle, line width=.05cm, draw, myBlue, minimum width=.75cm, minimum height=.75cm] at (0,2) {};
    \node (feats3) [rectangle, line width=.05cm, draw, BrickRed, minimum width=.75cm, minimum height=.75cm] at (1,2) {};

    \node (input) [] at (0,0) {Input};
    
    %\draw[->] input -| feats1;
    \draw [->] (input) -- (feats1);
    \draw [->] (input) -- (feats2);
    \draw [->] (input) -- (feats3);
    \draw [->] (feats1) -- (classifier);
    \draw [->] (feats2) -- (classifier);
    \draw [->] (feats3) -- (classifier);
    \draw [->] ([yshift=.5cm]classifier.north) -- ([yshift=.75cm]classifier.north);

\end{tikzpicture}
        \caption{\texttt{Multilingual} $\;$ model.}
        \label{fig:multiling}
    \end{subfigure}
    \hspace{.3cm}
    \begin{subfigure}{.2\textwidth}
        \begin{tikzpicture}
    \path[use as bounding box] (-1.5,-.5) rectangle (1.5,4.5);

    %\draw [myBlue, line width=.05cm] (0,4) ellipse (.65cm and .35cm);
    \node (classifier21) [rectangle, line width=.05cm, draw, myBlue, minimum width=.5cm, minimum height=.5cm, rotate=45] at (0,4) {};

    \node (classifier) at (0,3.75) {};
    \node (feats1) [rectangle, line width=.05cm, draw, ForestGreen, minimum width=.75cm, minimum height=.75cm] at (-1,2.5) {};
    \node (feats2) [rectangle, line width=.05cm, draw, myBlue, minimum width=.75cm, minimum height=.75cm] at (0,2.5) {};
    \node (feats3) [rectangle, line width=.05cm, draw, BrickRed, minimum width=.75cm, minimum height=.75cm] at (1,2.5) {};

    \draw [line width=.05cm] (0,1) ellipse (.35cm and .35cm);
    \node (langId) [rectangle, minimum width=.65cm, minimum height=.65cm] at (0,1) {};
    
    \node (input) [] at (0,0) {Input};
    
    %\draw[->] input -| feats1;
    \draw [->, dashed] (langId) -- (feats1);
    \draw [->, dashed] (langId) -- (feats3);
    %\draw [->] (langId) -- (feats2);
    \draw [->] (input) -- (langId);
    \draw [->] (feats1) -- (classifier);
    \draw [->] (feats2) -- (classifier);
    \draw [->] (feats3) -- (classifier);
    \draw [->] (langId) .. controls ++(0.65, 1.25) and ++(.65, -1)  ..  (classifier);
    \draw [->] (input) .. controls ++(-0.65, 1) and ++(-.65, -1)  ..  (feats2);
    \draw [->] ([yshift=.5cm]classifier.north) -- ([yshift=.75cm]classifier.north);
    
\end{tikzpicture}
        \caption{\texttt{Language-aware} multilingual model.}
        \label{fig:langaware}
    \end{subfigure}
    \hspace{.3cm}
    \begin{subfigure}{.15\textwidth}
        \begin{tikzpicture}
    \path[use as bounding box] (0,-.5) rectangle (3.5,-3.5);

    \draw [ForestGreen, line width=.2cm] (0,0) -- (1,0);
    \node [] at(2,0) {\begin{minipage}{1.5cm}Lang. 1\end{minipage}};

    \draw [BrickRed, line width=.2cm] (0,-.5) -- (1,-.5);
    \node [] at(2,-.5) {\begin{minipage}{1.5cm}Lang. 2\end{minipage}};

    \draw [myBlue, line width=.2cm] (0,-1) -- (1,-1);
    \node [] at(2,-0.9) {\begin{minipage}{1.5cm}Lang.\end{minipage}};
    \node [] at(2,-1.2) {\begin{minipage}{1.5cm}agnostic\end{minipage}};

    \node [rectangle, draw, myBlue,  line width=.05cm, minimum width=.5cm, minimum height=.5cm] at (.25,-2) {};
    \node [] at(1.5,-1.8) {\begin{minipage}{1.5cm}Feature\end{minipage}};
    \node [] at(1.5,-2.1) {\begin{minipage}{1.5cm}extractor\end{minipage}};
    
    \draw [line width=.05cm] (.25,-2.75) ellipse (.25cm and .25cm);
    \node [] at(1.5,-2.60) {\begin{minipage}{1.5cm}Language\end{minipage}};
    \node [] at(1.5,-2.90) {\begin{minipage}{1.5cm}classifier\end{minipage}};

    \node (classifier21) [rectangle, line width=.05cm, draw, myBlue, minimum width=.4cm, minimum height=.4cm, rotate=45] at (.25,-3.5) {};
%\draw [myBlue,  line width=.05cm] (.25,-3.5) ellipse (.3cm and .15cm);
    \node [] at(2,-3.4) {\begin{minipage}{2.5cm}Random forest\end{minipage}};
    \node [] at(1.5,-3.7) {\begin{minipage}{1.5cm}classifier\end{minipage}};
\end{tikzpicture}
    \end{subfigure}}
    \caption{Overview of the different proposed variations of MoNoise. Dashed
lines mean that only one of the two paths is taken, decided by the language
identification. For model (a), there can be two versions, one with features
from Lang. 1 (shown here) and one based on Lang. 2.}
    \label{fig:models}
\end{figure*}
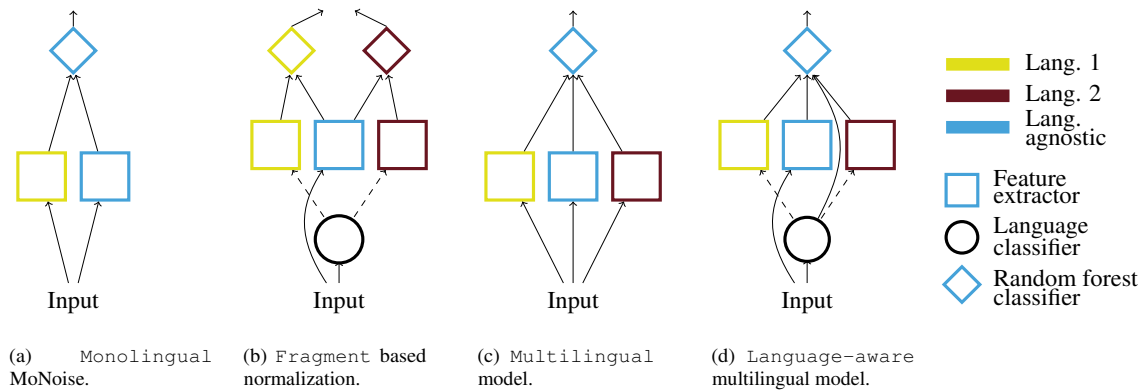

In this section we describe the models used for word-level language
identification (\ref{sec:method:lang}), lexical normalization
(\ref{sec:monoise}) and POS tagging (\ref{sec:pos}). 

\subsection{Word-level language identification}
\label{sec:method:lang}
We treat language identification as a sequence labeling task where the label of
each word is a language ID. We evaluate three sequence labeling libraries: 
\begin{inparaenum}[1)]
\item MarMoT~\cite{mueller-etal-2013-efficient}, a higher-order conditional
random fields tagger 
\item Bilty~\cite{plank-etal-2016}, a BiLSTM tagger, also incorporating
character level information
\item a BERT-based~\cite{devlin-etal-2019-bert} tagger named
MaChAmp~\cite{vandergoot2021massive}.
\end{inparaenum}
For Bilty, we project polyglot embeddings~\cite{polyglot:2013:ACL-CoNLL} of
each language of the language pairs to the same space using
MUSE~\cite{conneau2017word}, whereas for MaChAmp, we use multilingual
BERT.\footnote{multi\_cased\_L-12\_H-768\_A-12} We use the default settings for
all toolkits.

\subsection{Normalization}
We choose to use MoNoise~\cite{van-der-goot-2019-monoise} as a baseline and
starting point for our proposed models for two main reasons: 
\begin{inparaenum}[1)]
\item Normalization annotation for code-switched data is scarce, and MoNoise is
specifically strong in low-resource setups because of its dependence on
external resources (generated from raw data); 
\item It is the only normalization model that has shown to be effective in
multiple languages. 
\end{inparaenum}
Below we first introduce the standard monolingual MoNoise model, and then all
the proposed extensions which are focused on code-switched data.  A schematic
overview of all models is shown in Figure~\ref{fig:models}.

\paragraph{Monolingual} (Figure~\ref{fig:monoise})
\label{sec:monoise}
\noindent MoNoise consists of two parts, a candidate generation step and a
candidate ranking step. For the generation of candidates, a spelling correction
system (Aspell),\footnote{\url{www.aspell.net}} word embeddings and a
dictionary based on the training data are used.  Features from these modules
are then supplemented with n-gram probabilities based on Wikipedia and Twitter
data and other features indicating whether a word is present in the Aspell
dictionary, whether it contains an alphabetical character, the length of a
candidate compared to the original word,  and whether it starts with a capital.
For the novel proposed models, we will split up the features based on whether
they require \textit{language-specific} resources (spelling correction, word
embeddings and n-grams features; yellow and red in Figure~\ref{fig:models}), or
whether they are \textit{language-agnostic} (all other features; blue in
Figure~\ref{fig:models}).  For the ranking of the candidates a random forest
classifier~\cite{breiman2001random} is used, which predicts the probability
whether a candidate is correct.  An obvious disadvantage when applying
monolingual MoNoise on CS-data is that many features are language-specific
(e.g. spelling correction, word embeddings, n-grams), which is sub-optimal for
tokens from another language.  Since our datasets and evaluation include
capitals, we use the version of MoNoise including capitalization
handling~\cite{goot-etal-2020-norm}.

\paragraph{Fragments} (Figure~\ref{fig:fragment})
\noindent The baseline model has the deficiency that it has the
language-specific features only for one language, while normalizing texts for
two languages.  An intuitive way of improving this model would be to split up
the input data into monolingual fragments, and train two separate monolingual
models. The fragments are created by splitting the data on every CS point,
where words with the \un label are converted to the label of the previous word.
This setup has the advantage that the normalization model itself does not need
any adaptation, and it can thus be used with any normalization model. The
disadvantages are that it is dependent on a language label, two separate
classifiers have to be trained and the context is interrupted.

\paragraph{Multilingual} (Figure~\ref{fig:multiling})
\noindent Instead of using two separate random forest classifiers, we can
exploit both feature sets simultaneously in one classifier. This means that for
every language-specific feature, we now have two features.  In this setup, the
model is not explicitly informed about the language of input words, however,
some of the features (especially n-gram probabilities) will have a very high
correlation with this information.  This model has the advantage that only one
classifier has to be trained, and no language labels are necessary. It has the
disadvantage that it uses more features for the classifier compared to the
\texttt{Monolingual} and \texttt{Fragments} models, which increases the
complexity of the classification.

\paragraph{Language-aware} (Figure~\ref{fig:langaware})
\noindent Some of the language-specific features of the \texttt{Multilingual}
model will be rather superfluous for words in the other language. For example,
it will search for Turkish words in German word embeddings, and also use n-gram
counts based on the German Wikipedia. To avoid this, we can use only one copy
of each language-specific feature, and generate them based on the language
label (the same language labels as in the \texttt{Fragments} model are used).
More concretely, this means that for a German word, we will generate uni-gram
probabilities based on German data, whereas for Turkish we will use Turkish
data; these are then modeled as one feature in the model. On top of this, we
also add a feature that indicates which language a word belongs to.  There
might be some mismatches in the importance of features because different data
sources and languages are used. Because the language label is known, and a
random forest classifier can model feature interactions
intrinsically~\cite{breiman2001random}, these mismatches should not be
problematic.  This model has the advantage that the number of features stays
almost the same as in the \texttt{Monolingual} model (+1, the language ID), but
a disadvantage is that it requires language labels. 

\subsection{POS tagging}
\label{sec:pos}

For POS tagging, we examine the same three sequence labeling systems as used
for language identification (Section~\ref{sec:method:lang}): MarMoT, Bilty and
MaChAmp.  For each normalization setting, we normalize the input data, and use
this normalized text as input for the POS tagger, which is trained on canonical
data.

\section{Evaluation}

\begin{table}
\centering
\begin{tabular}{l r r}
\toprule
Model & \iden & \trde \\
\midrule
MarMoT & 92.71 & 92.91 \\
Bilty & $^*$93.81 & $^*$94.31\\
MaChAmp & $^*$95.17 & $^*$95.67 \\
\bottomrule
\end{tabular}
\caption{Word level accuracies for language identification (10-fold).}
\label{tab:langId}
\end{table}

In this section we evaluate each of the three sub-tasks (LID, normalization,
POS), where for the latter two we also examine the effect of exploiting the
prediction of the previous tasks. Unless mentioned otherwise, we report the
results of 10-fold cross-validation on the training split of the data. For all
experiments, we use a paired bootstrap test on the sentence level with 1,000
samples to test significance. For all results, we order the models by the
complexity of the implementation as compared to MoNoise (first fragments, as
the original model can be used as a black box, then multilingual because it
does not need a language classifier, and finally the language-aware model).  An
$^*$ next to results denotes a significant difference for $p < 0.05$, of a
model always as compared to the previous model (corresponding to the previous
column in Table \ref{tab:posResults}, the previous row in other tables) for the
same data.  

\subsection{Language identification}
\label{sec:langidEval}

Results for the language identification task are reported in
Table~\ref{tab:langId}. Unsurprisingly, the performances are in line with the
chronological order of the introduction of the systems, and their computational
complexity. It should be noted that for MaChAmp we used  pre-trained embeddings
which were trained on the largest amount of external data. When inspecting the
performance per language label, we saw that the `{\un}specified' is by far the
most difficult. Even though this class contains punctuation, it also contains
many harder cases, where a word belongs to any language other than Lang1 and
Lang2, or when the annotator is uncertain.
~\newcite{barik-etal-2019-normalization} use a conditional random fields
classifier with a variety of features for this task, and report 90.11 accuracy
for the full \iden dataset in a 5-fold cross-validation setting. Which, despite
differences in data splits, confirms that our results are competitive.

\subsection{Normalization}

\begin{table}
\centering
          \begin{tabular}{l | r r r | r r r}
        \toprule
Model & Id-En & Tr-De \\
\midrule
LAI & 73.24 & 74.03\\
MFR & $^*$88.35 & $^*$78.57\\
Monolingual-lang1 (Tr/Id) & $^*$94.76 & $^*$79.81\\
Monolingual-lang2 (De/En) & $^*$94.31 & 80.58\\
\midrule
Fragments  & 94.73 & $^*$81.24\\
Multilingual & 94.84 & $^*$81.74\\
Language-aware & 94.79 & 81.68\\
\bottomrule
    \end{tabular}
    \caption{Normalization performance of the baselines and the proposed models (10-fold accuracy). For the models dependent on language labels, we used the labels predicted by MaChAmp.}
    \label{tab:results}
\end{table}

For lexical normalization, a wide variety of evaluation metrics is used in the
literature, ranging from accuracy~\cite{han-baldwin-2011-lexical}, F1
score~\cite{baldwin-etal-2015-shared} and precision over out-of-vocabulary
words~\cite{alegria2013introduccion}, to CER and BLUE
score~\cite{ljubevsic2016normalising}. Because the word order is fixed in our
task, and to ease interpretation of the results, we opt to use simple accuracy
on the word level, where we consider all words (i.e., also the %words which are
not normalized).  unnormalized words).

To interpret the scores, we include three baselines: 
\begin{inparaenum}[1)]
\item leave-as-is (LAI), which always outputs the original word, i.e.\ its
accuracy is equivalent to the percentage of words that are not normalized 
\item most-frequent-replacement (MFR), which uses the most frequent replacement
from the training data for each word
\item monolingual MoNoise, which can be trained on either of the languages
within a language pair (two models).
\end{inparaenum}

Results for the different models are compared in Table~\ref{tab:results}.  For
the \iden dataset, the differences between all proposed models are small and
not significant. Even the monolingual models perform remarkably well, and only
small gains are observable when using the multilingual model.  We also compared
our results to~\newcite{barik-etal-2019-normalization}, using their evaluation
metric as their model/output was not available. The metric is
non-deterministic, as it uses accuracy over unique OOV words.\footnote{Which
can be normalized differently dependent on context, we confirmed this with the
authors.} Nevertheless, our average estimated result for \texttt{Multilingual}
is 69.83 for this metric, outperforming their score of 68.50.

For the \trde dataset, the scores are generally lower, indicating that this
dataset (and perhaps language pair) is more difficult.  Especially now, we can
observe that the code-switched adaptations lead to substantially higher scores.
To our surprise, \texttt{Multilingual} and \texttt{Language-aware} perform on
par, even though the multilingual model does not rely on language labels.
\texttt{Fragments} performs significantly worse. This leads to the conclusion
that language labels are not directly beneficial for lexical normalization (in
this setup). In general, the performances are in a similar range as for
monolingual
datasets~\cite{van-der-goot-2019-monoise}.\footnote{\newcite{van-der-goot-2019-monoise}
used error reduction rate as main evaluation metric, for which the multilingual
model would score 80.72 (\iden) and 30.42 (\trde). The reported scores on
monolingual datasets are 77.09 for En and 28.94 for Tr.}

\paragraph{Model behavior}
Besides the metrics reported in the table, we also examined precision and
recall. Precision is generally much higher (1.1 to 3 times, see
Appendix~\ref{app:precRec}) than recall especially for \trde, which is in line
with previous observations~\cite{van-der-goot-2019-monoise}. This means that
the model is conservative and only replaces cases for which it is rather
certain, which arguably is a desirable behavior. 

\begin{table}
\centering
\resizebox{\columnwidth}{!}{
\begin{tabular}{l | r r}
        \toprule
Model & Id-En & Tr-De \\
\midrule
Fragments (MarMoT) & 94.66 & 80.77\\
Fragments (Bilty) & $^*$94.71 & $^*$80.89\\
Fragments (MaChAmp) & 94.73 & $^*$81.24\\
 \hdashline
 Fragments (Gold) & $^*$94.81 & 81.71\\
\midrule
Language-aware (MarMoT) & 94.74 & 81.24\\
Language-aware (Bilty) & 94.76 & 81.57\\
Language-aware (MaChAmp) & 94.79 & 81.68\\
 \hdashline
Language-aware (Gold) & 94.90 & 82.18\\
\bottomrule
    \end{tabular}}
    \caption{Effect of different language predictions on normalization models
(10-fold accuracy).}
    \label{tab:langIdNorm}
\end{table}

\paragraph{Effect of language predictions}
To evaluate the effect of the language predictions, we run both the
\texttt{Fragments} and the \texttt{Language-aware} models with all language
predictions from Section~\ref{sec:langidEval} as well as the gold language
labels. The results (Table~\ref{tab:langIdNorm}) show that the performance of
the language identification has a positive effect on the normalization
performance. Although it is not significant in most cases, it should be noted
that significance is only tested compared to the previous model.

\begin{table*}
\centering
\begin{tabular}{l | r r r r}
\toprule
 Model & LAI & Multiling. & Lang.-aware & Gold \\
 \midrule
MarMoT--POS & 61.92 & $^*$65.50 & 65.47 & $^*$69.14 \\
Bilty--POS  & $^*$65.23 & $^*$67.99 & $^*$68.26 & $^*$72.04 \\
MaChAmp--POS  & 65.60 & $^*$68.25 & 68.13 & $^*$71.27 \\
\bottomrule
\end{tabular}
\caption{Accuracies for \trde POS tagging, using a variety of normalization
strategies.}
\label{tab:posResults}
\end{table*}

\paragraph{Language labels}
Looking at the normalization performance breakdown on language labels shows
that the gains of our proposed models are consistently smaller on Indonesian
and Turkish compared to respectively English and German (see
Appendix~\ref{app:perLang} for full results). This was to be expected, as for
these languages the model has less external data (Section~\ref{sec:rawData})
and while the model was originally not evaluated for Indonesian, Turkish had
the lowest performance in~\newcite{van-der-goot-2019-monoise}.

\paragraph{Qualitative analysis}
Both \texttt{Multilingual} and \texttt{Language-aware} correct most frequent
normalization mistakes well. This means for \iden, abbreviations (\textit{yg}
$\mapsto$ \textit{yang} `which is' in \ind), slang words (\textit{gw} $\mapsto$
\textit{saya}, \textit{gue} $\mapsto$ \textit{saya} `I' in \ind), phonetic
spelling (\textit{kalo} $\mapsto$ \textit{kalau} `if' in \ind); and for \trde
emoticons, restoring Turkish-specific characters, restoring  vowels
(\textit{cnm} $\mapsto$ \textit{canım} `my dear' in \tr), and punctuation
replacements. On the \iden data, however, there is a higher number of these
frequent replacements compared to the \trde dataset, which explains the high
scores and small variability for \iden in Table~\ref{tab:results}
and~\ref{tab:langIdNorm}.

For the \trde dataset, the most common mistakes include: not correcting
capitalization in the beginning of a sentence, merging of words, monolingual
ambiguous cases depending on context (\textit{mi} $\mapsto$ [\textit{mi},
\textit{mı}], question clitics in \tr), and tokenization and punctuation
mistakes (?:D $\mapsto$? :D). In comparison, for the \iden  dataset, the models
make rather different errors: in-vocabulary words which should be normalized
are left as is (\textit{kaya}  $\mapsto$ \textit{seperti}, \textit{usah}
$\mapsto$ \textit{perlu}), normalizations which are lexically very distant are
not found (\textit{lw} $\mapsto$ \textit{kamu}), and English contractions are
often not replaced (\textit{isnt} $\mapsto$ \textit{is not}).  Error analysis
on the \iden dataset revealed that correction of capitalization was annotated
inconsistently.  However, because in most cases the normalization was
lowercased, this did not have a large effect on performance. 

Interestingly, \texttt{Language-aware} is better in correcting words that exist
in both languages. For instance, \textit{ne} is the informal form of
\textit{eine} `a/one' in German, and also means `what' in Turkish. The dataset
annotations expect the \textit{ne} $\mapsto$ \textit{eine} normalization. While
\texttt{Multilingual} fails to do so,  \texttt{Language-aware} corrects them.
We believe language IDs play a positive role here in defining the context, and
although in general both models perform on par, if a dataset contains many such
ambiguous words, \texttt{Language-aware} could be preferable.

\subsection{POS tagging}
\label{sec:evalPos}
For POS tagging, we only look at \trde as \iden is not annotated with POS tags.
We employ a pipeline approach; we first normalize our training data in a
10-fold setting, and then apply the tagger on this normalized data.  The
taggers are trained on a shuffled concatenation of the Turkish-IMST
\cite{sulubacak2016} and German-GSD \cite{mcdonald2013} datasets of UD version
2.5~\cite{nivre-etal-2020-universal}.  Now that none of the CS data is used
during training, 10-fold cross-validation is not necessary.  We directly apply
the taggers on the full training data. This way the exact same data split is
used for evaluation as in the 10-fold setting in the previous sections.  Even
though we have POS tags available for the gold normalization
(Section~\ref{sec:trdedata}), we do not have gold tags for predicted
normalization, and to keep the comparison fair we evaluate using the \tokanon
POS tags.  When a word is split or merged, we use the alignment and check
whether the correct tag is present. In other words: we select one tag based on
an oracle selection.\footnote{It should be noted that this makes splitting
beneficial, and this metric can easily be tricked by splitting every token so
it should be used with caution. However, our proposed normalization models have
a low rate of splitting (114 versus 398 in gold) and merging is not handled at
all.}

Results in Table~\ref{tab:posResults} show that, surprisingly, Bilty performs
competitive to MaChAmp across most settings. Considering the differences
between the normalization models, the \texttt{Multilingual} model and the
\texttt{Language-aware} model perform on par, but there is still a marginal gap
compared to the gold normalization.

We also analyzed the confusion matrices of the POS tagger, the full analysis
can be found in Appendix \ref{app:posConfusion}, we will shortly summarize
findings here.  1) Bilty is mainly outperforming MaChAmp in gold due to better
recognition of symbols (emojis), 2) Bilty is more sensitive to different
normalization strategies, whereas for MaChAmp the differences between them are
minimal, 3) Performance on nouns improves a lot after normalization, especially
for German (due to corrected capitalization of nouns), 4) The second POS tag
which improved most are verbs, investigation showed that this is mainly because
Turkish-specific characters are replaced by their ASCII counterparts, which
helps the tagger assign the correct POS.

\subsection{Test data}
\label{sec:test}
On the test data we take both the `no normalization' and the best baseline
(which are monolingual Indonesian for \iden and monolingual German for \trde),
and compare these to our best two proposed normalization models.  The results
in Table~\ref{tab:test.norm.pos} show that, parallel to 10-fold
cross-validation results (Table~\ref{tab:results}), \texttt{Multilingual}  and
\texttt{Language-aware} scores are similar and their difference is
insignificant for both datasets. This leads to the conclusion that
\texttt{Multilingual} is the most elegant model, as it is not dependent on
language labels. On the \trde dataset the proposed models are clearly
outperforming the baselines. However, on the \iden dataset the differences are
small (and not significant) between the monolingual model and both of our
proposed models. 

For \trde, we take the test set normalized by systems in the second column of
Table~\ref{tab:test.norm.pos} and apply MaChAmp for POS tagging. The results in
the third column show that the POS tagger follows the trend in normalization
scores, and performs slightly better when using the multilingual model, beating
the LAI baseline (i.e. not using normalization) with 5.4\% relative
improvement.

\begin{table}
 \resizebox{\linewidth}{!}{
\begin{tabular}{l | r r |  r}
        \toprule
         & \multicolumn{2}{c}{Normalization} & POS\\
Model & Id-En & Tr-De & Tr-De \\
\midrule
LAI             & 74.03 & 67.02 & 60.77\\
Monolingual (Id/De) & $^*$94.62 & 76.33 & $^*$63.47\\
\midrule
Multilingual    & 94.27 & $^*$78.28 & $^*$64.06 \\
Language-aware  & 94.32 & 77.83 & $^*$63.92\\
\hdashline
Gold            & $^*$100.00 & $^*$100.00& $^*$67.75\\
\bottomrule
    \end{tabular}
}    
    \caption{Normalization and POS tagging accuracies on test data. The POS
tagging model is the same MaChAmp model for all results, only the normalization
strategy for the input changes.}
    \label{tab:test.norm.pos}
\end{table}

\section{Conclusion}
Code-switching provides many challenges for NLP systems.  In this work we
attempt to overcome some of these challenges by normalizing the data, and
evaluating the downstream effect of this for POS tagging.  For evaluation we
use an Indonesian-English dataset~\cite{barik-etal-2019-normalization} as well
as a German-Turkish dataset~\cite{cetinoglu-2016-turkish}, for which we
provided novel normalization layers and adapted existing LID and POS
annotation.

We proposed three different models to normalize CS data. The two
best-performing models are \texttt{Language-aware} and \texttt{Multilingual}.
The first model exploits language labels, to identify for which language to
generate features, whereas the second model combines features for both
languages. The differences in performance between these two systems was not
significant for any of the 10-fold experiments nor on the test data, so in most
cases the multilingual model would be preferable, as it has no dependence on
language labels.

We showed that normalizing the input before POS tagging results in
significantly higher POS accuracies for CS data. Gold normalization experiments
showed that there is still room for improvement for normalization models to
help POS tagging.

An interesting property of the proposed model is that it does not have to be
trained on intrasentential CS data. In fact, it can be trained on a mix of two
monolingual datasets, thereby handling many more language pairs.  We hope to
evaluate this setup if resources (i.e., normalization test data for a CS
language pair, and monolingual normalization training data for both languages)
become available.

\section*{Acknowledgements}
We would like to thank Barbara Plank, Alan Ramponi, Marija Stepanovic, and
Agnieszka Falenska, for feedback on early drafts. We also thank Manuel Mager
and Sevde Ceylan with their help on \trde data alignment and Anab Maulana Barik
for sharing the \iden data. The first author is partially funded by an Amazon
Research Award. The second author is funded by DFG via project CE 326/1-1
``Computational Structural Analysis of German-Turkish Code-Switching'' (SAGT).

\bibliographystyle{acl_natbib}
\bibliography{papers}

\clearpage

\section*{Appendix}
\appendix
\section{Breakdown of performance per language}
\label{app:perLang}

Table~\ref{tab:languagesNorm} show the accuracy of all the proposed models per
language. The LAI scores show that most of the normalization replacements are
necessary for \ind and \DE. Interestingly, performance of the last two models
is highest on respectively \en and \DE, which is probably due to the original
model being developed mostly with a focus European languages.

\begin{table}
\centering
 \resizebox{\linewidth}{!}{
\begin{tabular}{l | r r r r}
\toprule
Model & \ind & \en & \tr & \DE \\
\midrule
LAI & 66.92 & 71.33 & 70.21 & 66.92 \\
MFR & 87.58 & 88.10 & 76.26 & 69.55 \\
Monoling1 (tr/id) & 92.71 & 95.82 & 78.24 & 70.17 \\
Monoling2 (de/en) & 91.78 & 95.75 & 76.81 & 77.50 \\
Frags & 84.71 & 82.03 & 70.10 & 63.92 \\
Multiling. & 92.91 & 95.80 & 78.27 & 78.73 \\
Lang-aware & 92.77 & 95.78 & 78.10 & 79.32 \\
\bottomrule
\end{tabular}}
\caption{Normalization accuracies per language (with gold language labels)}
\label{tab:languagesNorm}
\end{table}

\section{Precision and recall}
\label{app:precRec}
Table~\ref{tab:precRec} show the precision and recall of all models on both
datasets. LAI has 0.0 on all metrics, because it never finds a correct
normalization.

\begin{table}
\centering
 \resizebox{\linewidth}{!}{
\begin{tabular}{l | r r r r}
\toprule
 & \multicolumn{2}{c}{\iden} & \multicolumn{2}{c}{\trde} \\
Model & recall & precision & recall & precision \\
\midrule
LAI & 0.00 & 0.00 & 0.00 & 0.00 \\
MFR & 57.50 & 98.20 & 21.42 & 84.39 \\
Monoling1 (tr/id) & 82.15 & 97.96 & 25.55 & 88.50 \\
Monoling2 (de/en) & 80.35 & 98.03 & 29.49 & 87.32 \\
Frags & 82.05 & 97.88 & 31.15 & 90.21 \\
Multiling. & 82.51 & 97.87 & 32.98 & 90.85 \\
Lang-aware & 82.21 & 97.98 & 32.95 & 90.34 \\
\bottomrule
\end{tabular}}
\caption{precision and recall for both datasets, we follow the definitions
of~\cite{van-der-goot-2019-monoise}}
\label{tab:precRec}
\end{table}

\section{Confusions of POS taggers}
\label{app:posConfusion}
\begin{figure}
\begin{subfigure}{.5\textwidth}
    \includegraphics[width=.8\textwidth]{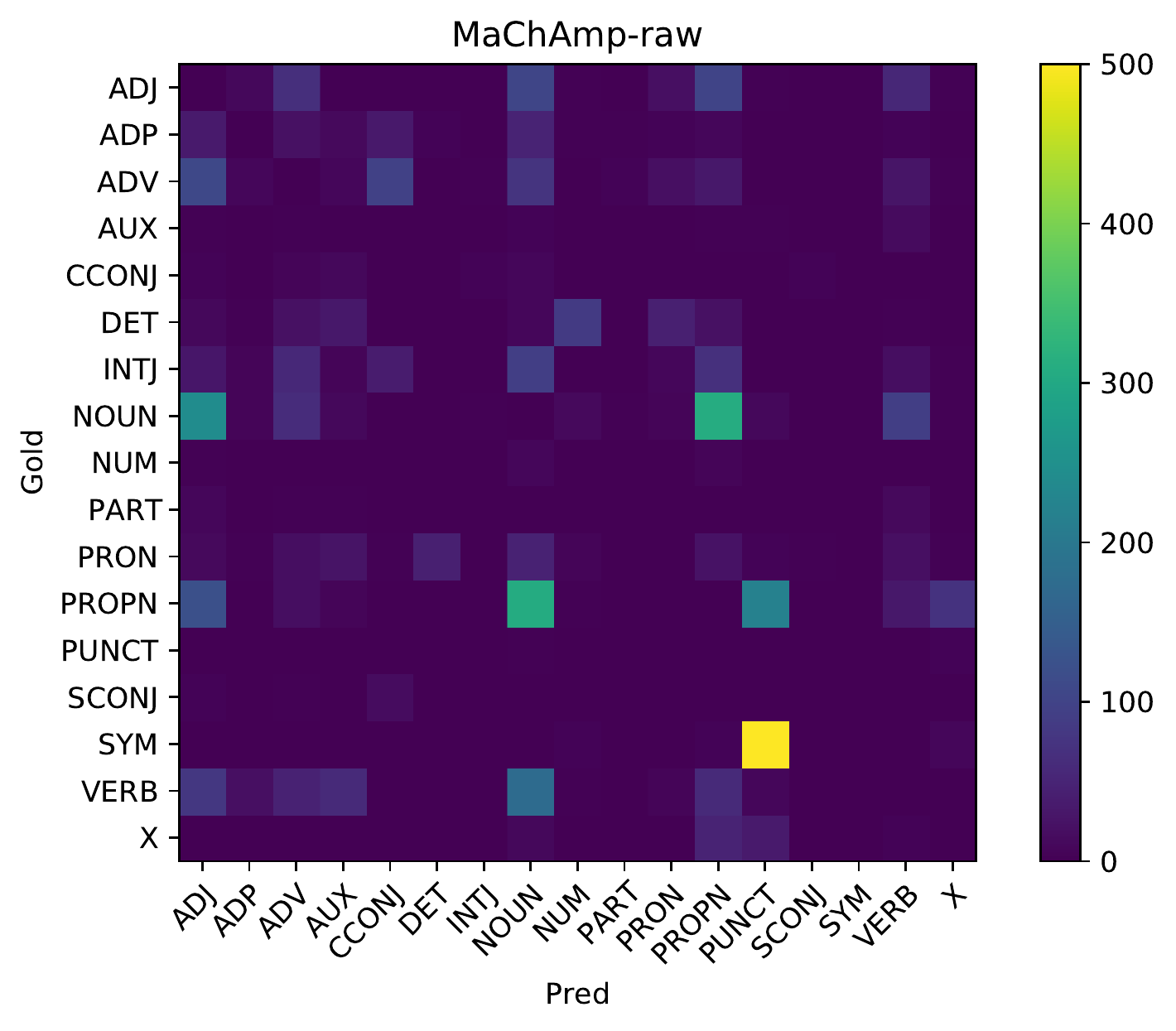}
\end{subfigure}
\begin{subfigure}{.5\textwidth}
    \includegraphics[width=.8\textwidth]{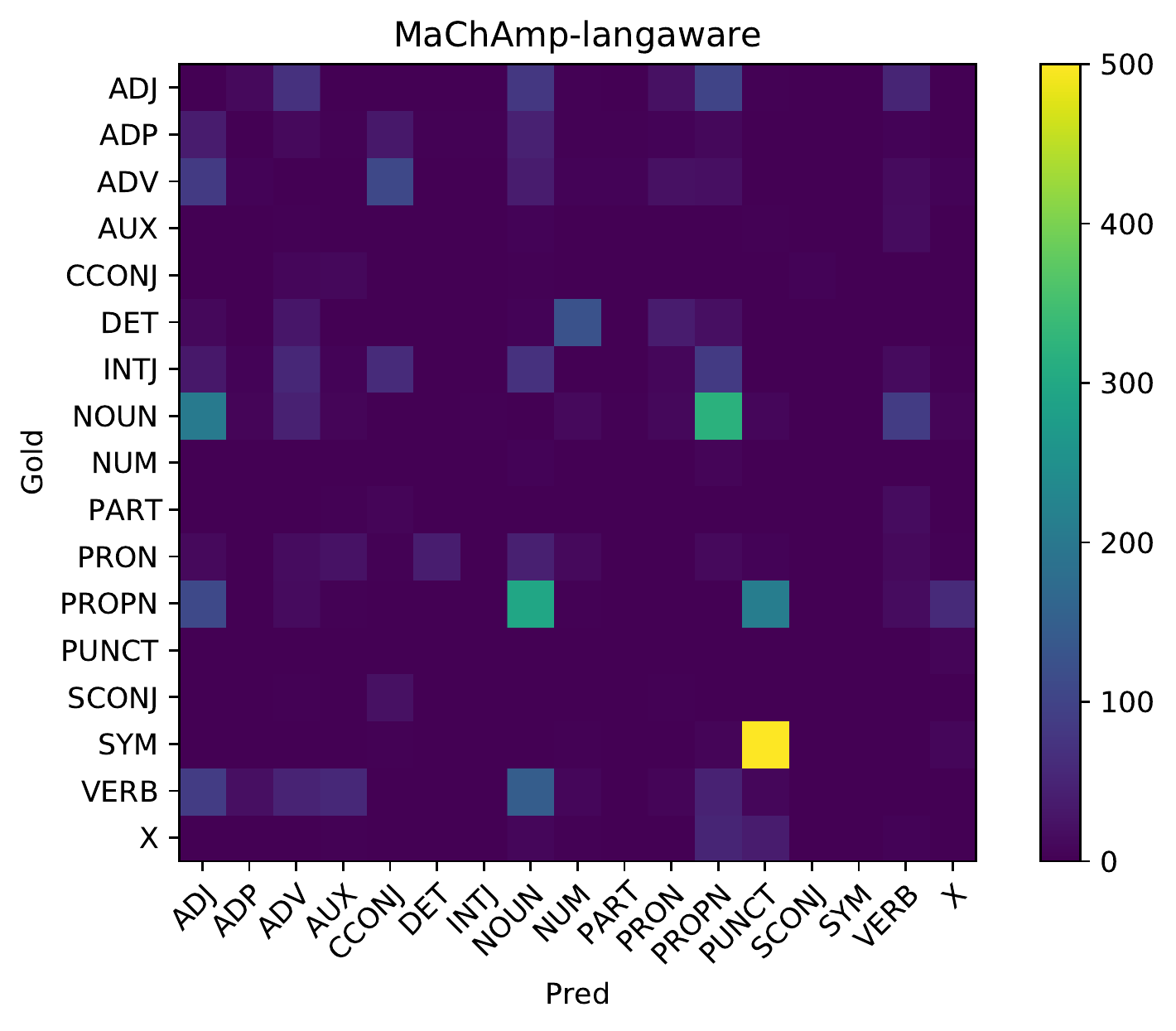}
\end{subfigure}

\begin{subfigure}{.5\textwidth}
    \includegraphics[width=.8\textwidth]{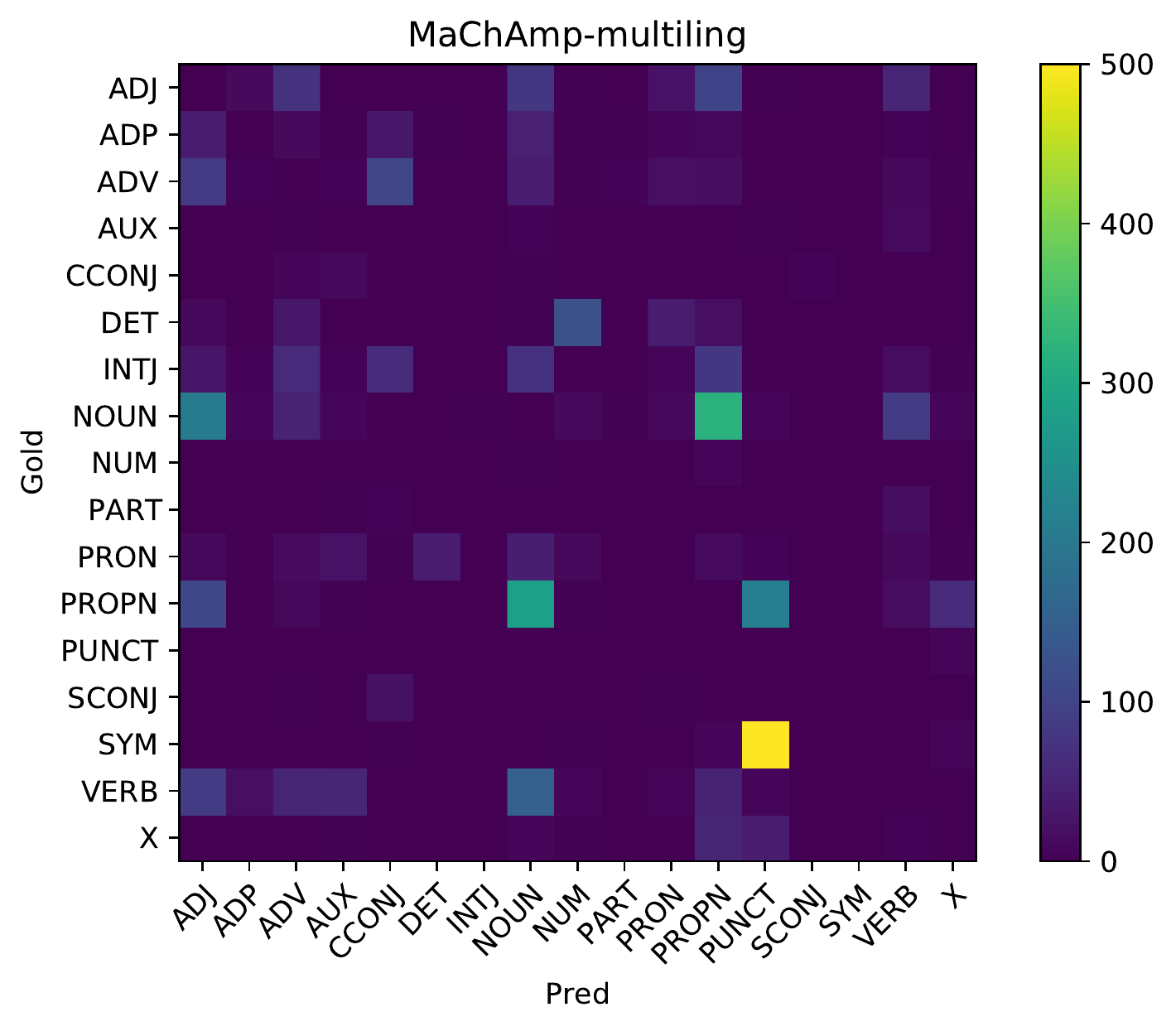}
\end{subfigure}
\begin{subfigure}{.5\textwidth}
    \includegraphics[width=.8\textwidth]{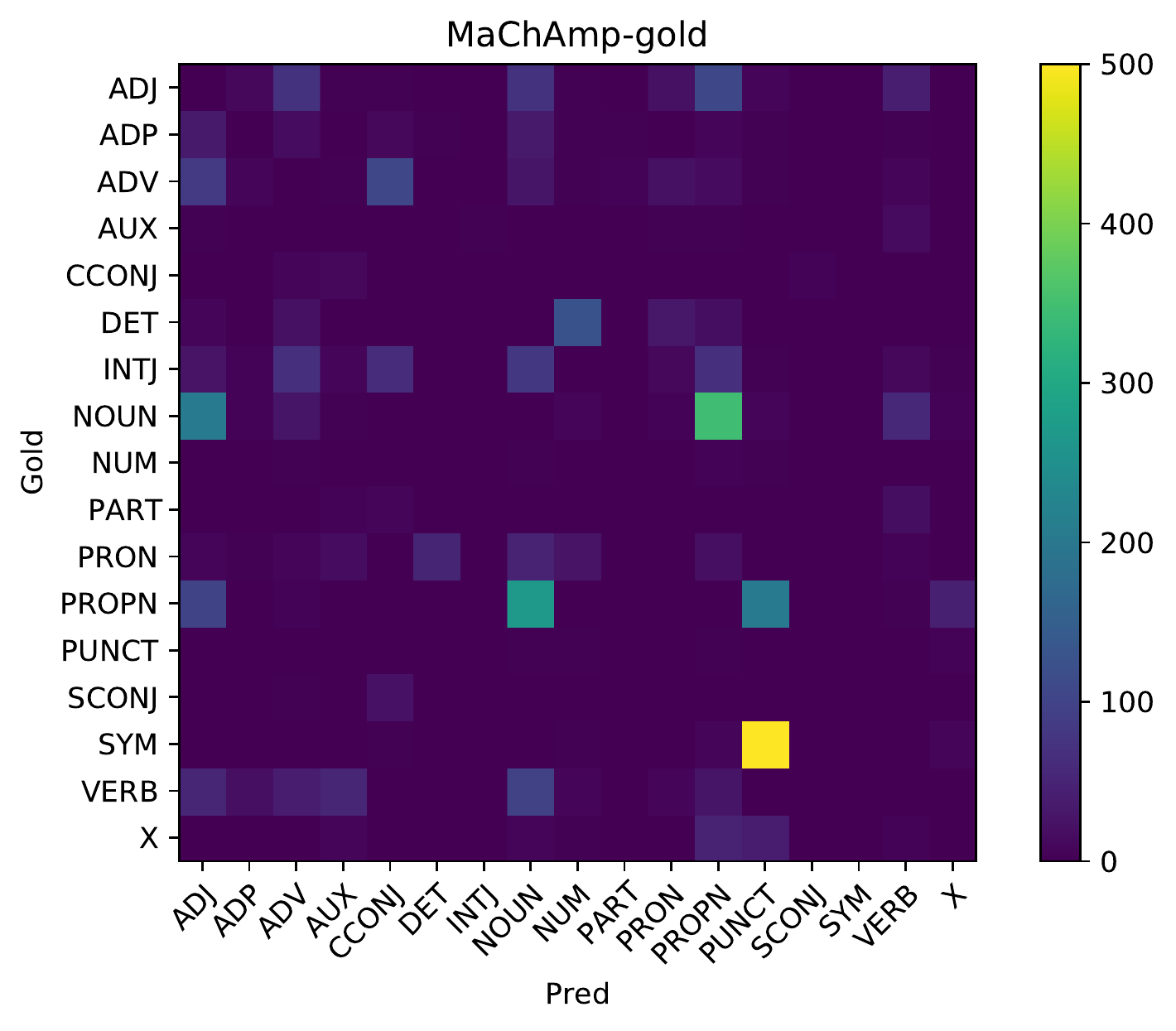}
\end{subfigure}
\caption{Confusion matrices for MaChAmp using a variety of normalization strategies}
\label{fig:confMachamp}
\end{figure}

\begin{figure}
\begin{subfigure}{.5\textwidth}
    \includegraphics[width=.8\textwidth]{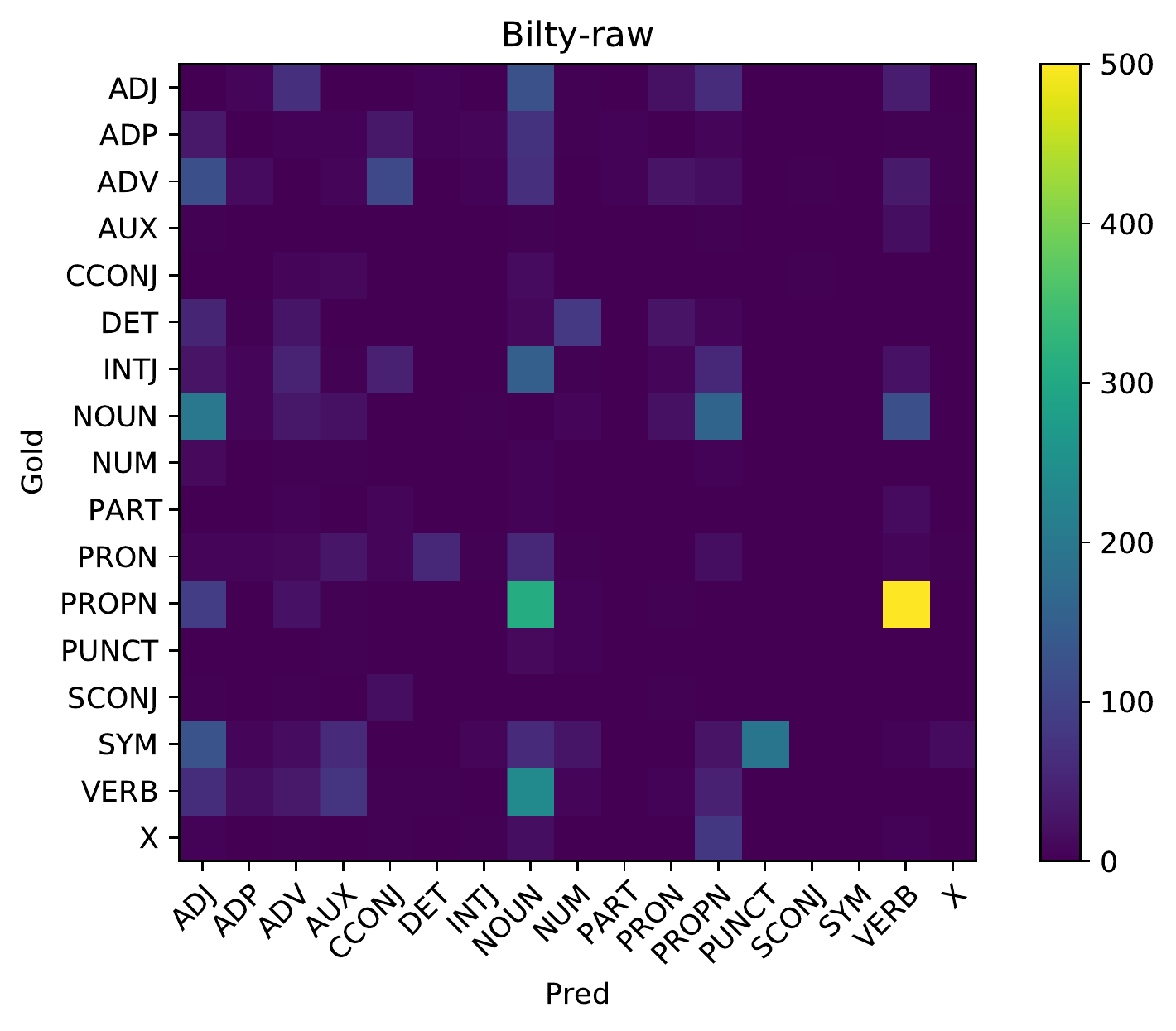}
\end{subfigure}
\begin{subfigure}{.5\textwidth}
    \includegraphics[width=.8\textwidth]{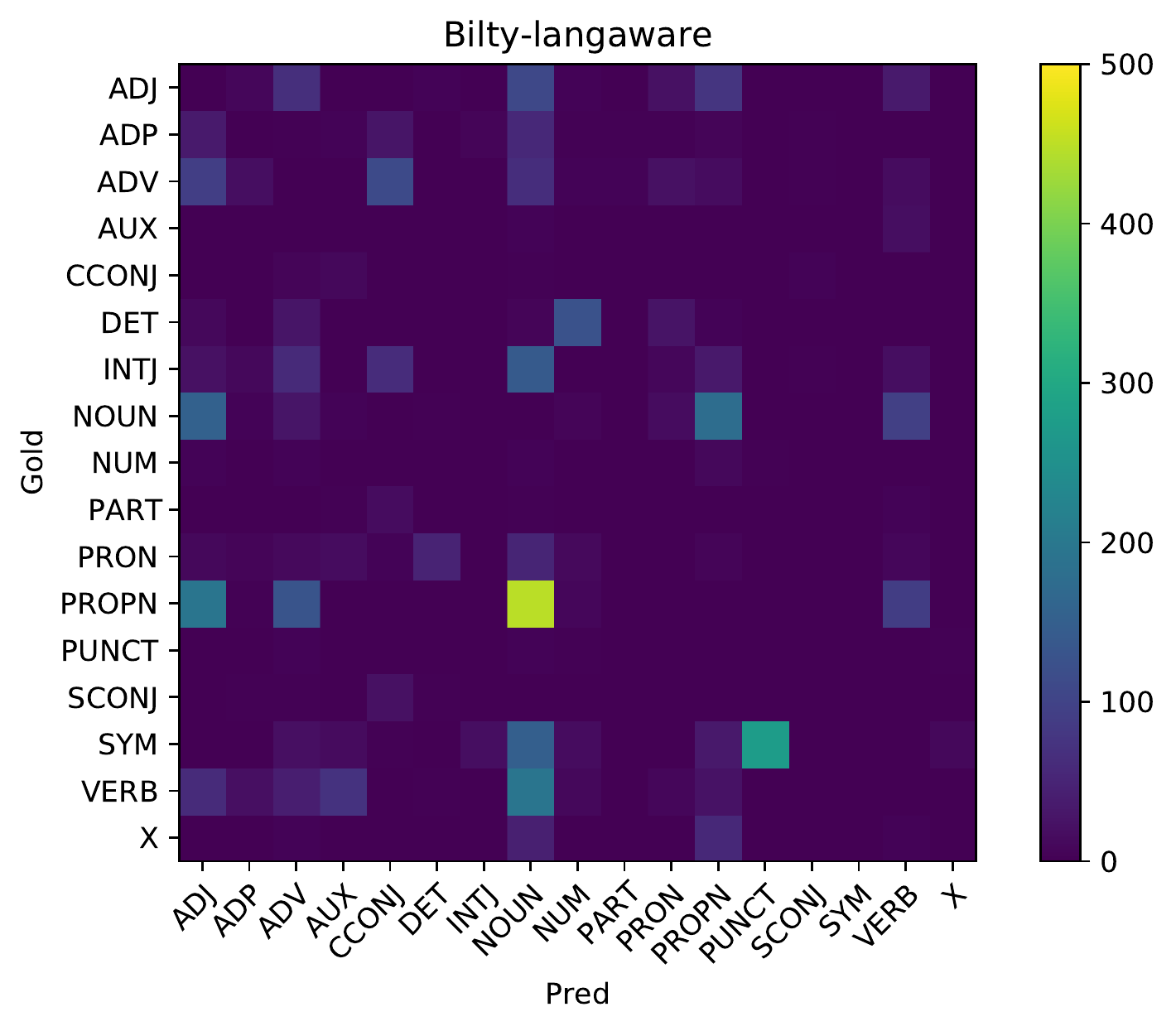}
\end{subfigure}

\begin{subfigure}{.5\textwidth}
    \includegraphics[width=.8\textwidth]{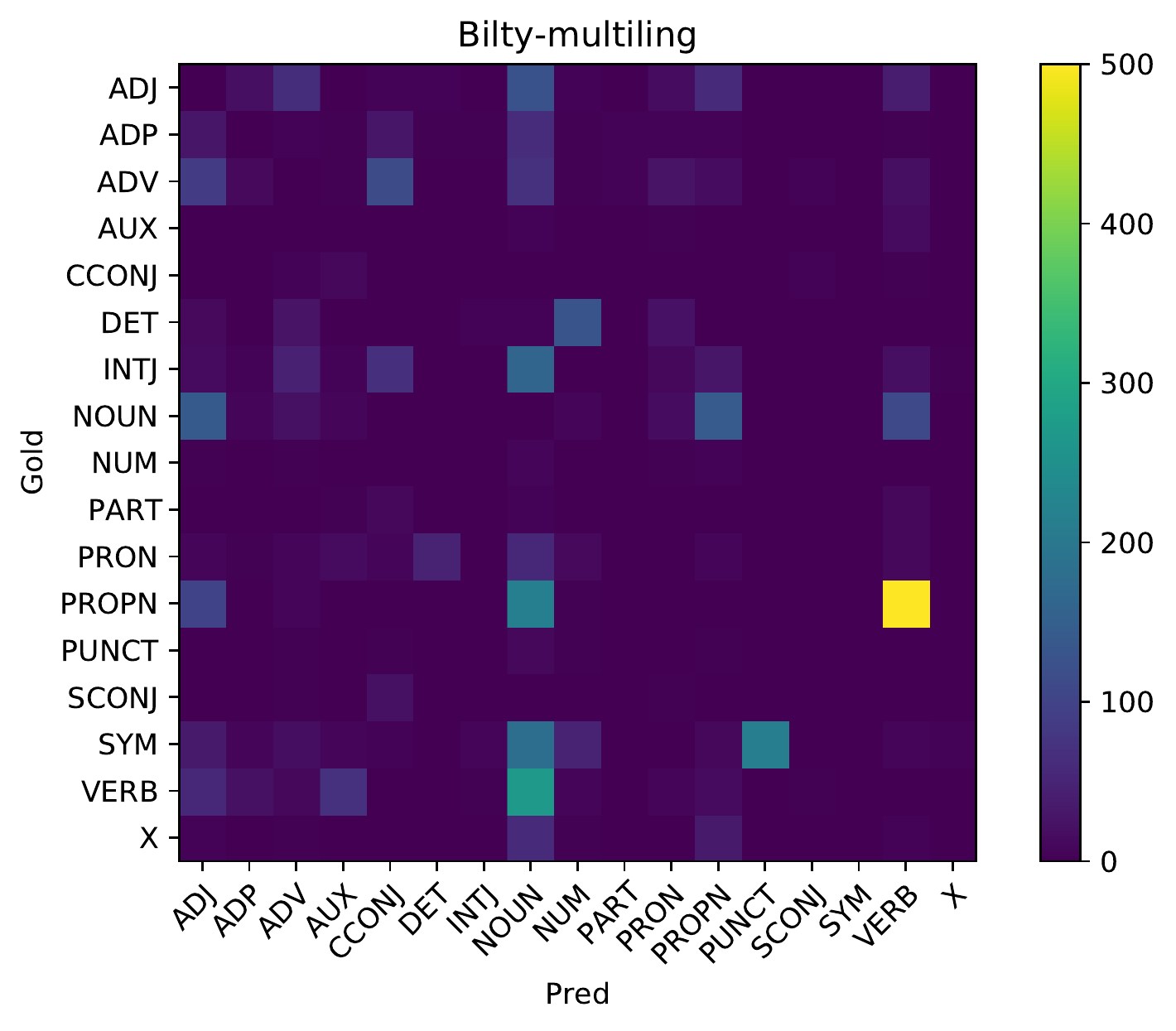}
\end{subfigure}
\begin{subfigure}{.5\textwidth}
    \includegraphics[width=.8\textwidth]{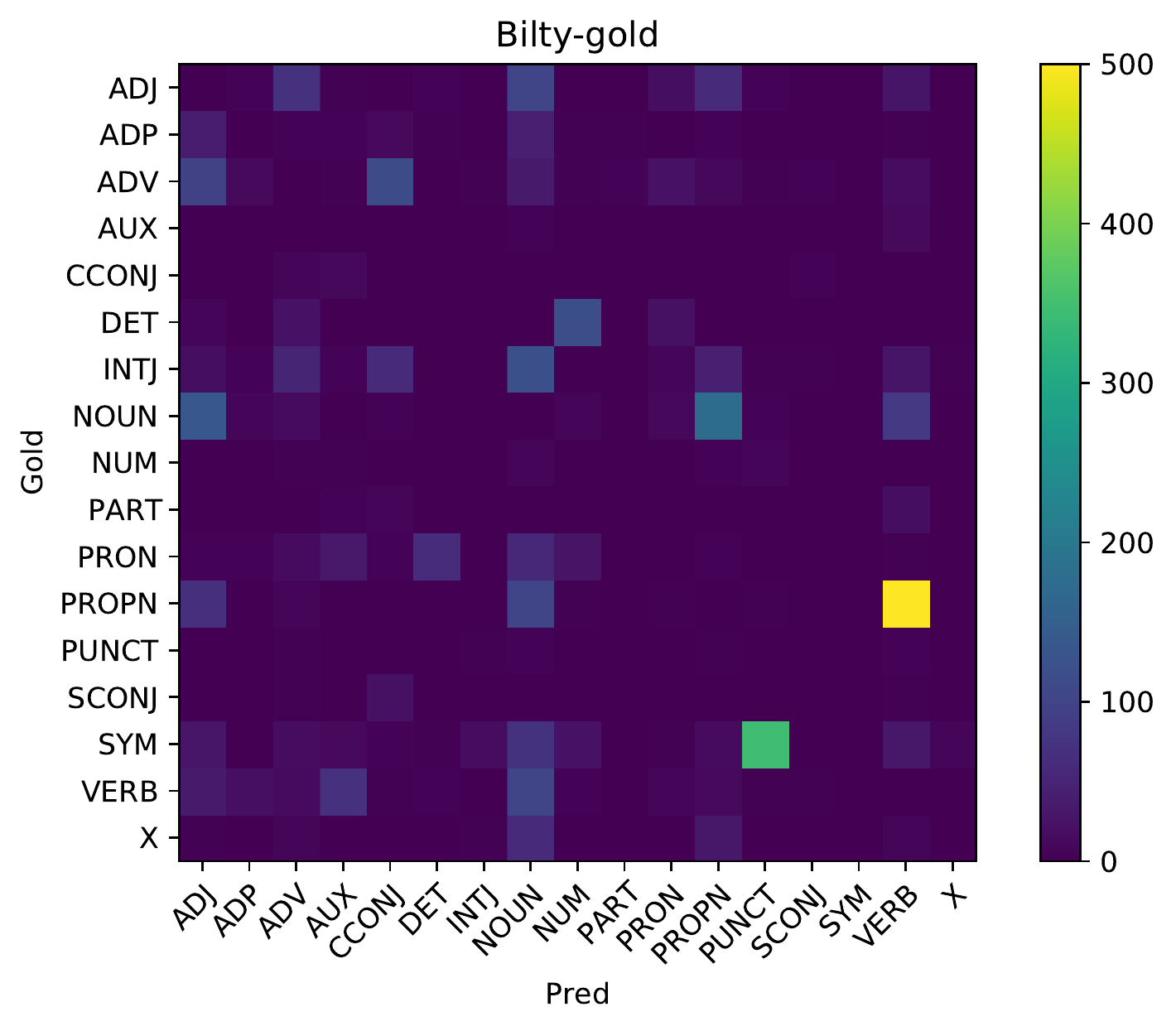}
\end{subfigure}
\caption{Confusion matrices for Bilty using a variety of normalization strategies}
\label{fig:confBilty}
\end{figure}

We conducted an analysis of POS tagging confusions for the setting described in
Section~\ref{sec:evalPos}.  In Table~\ref{tab:confMachamp} and
Table~\ref{tab:confBilty} the error frequencies of respectively MaChAmp and
Bilty are shown. The tables report the frequency of the top-10 most frequent
errors of the baseline (LAI), and the difference in counts observed using a
variety of normalization strategies.  In Figure~\ref{fig:confMachamp} and
Figure~\ref{fig:confBilty} the full confusion matrices for respectively MaChAmp
and Bilty are shown. For both of these analyses, we do not report the other
baselines, the fragment based model and the MarMot tagger, because performance
of these was inferior and this would make comparisons more complex.

\begin{table}
\centering
 \resizebox{\linewidth}{!}{
\begin{tabular}{l |r r r r}
\toprule
 & LAI & Multiling. & Lang-aware & Gold \\
\midrule
SYM-PUNCT & 529 & +2 & +2 & +34 \\
NOUN-PROPN & 310 & +10 & +10 & +36 \\
PROPN-NOUN & 307 & -24 & -11 & -39 \\
NOUN-ADJ & 244 & -40 & -38 & -40 \\
PROPN-PUNCT & 220 & -5 & -8 & -16 \\
VERB-NOUN & 174 & -21 & -27 & -78 \\
PROPN-ADJ & 122 & -13 & -12 & -20 \\
ADV-ADJ & 108 & -23 & -23 & -23 \\
ADJ-NOUN & 104 & -22 & -23 & -30 \\
ADJ-PROPN & 103 & -2 & +0 & +5 \\
\bottomrule
\end{tabular}}
\caption{10 most common POS tagging errors for LAI baseline, counted for all
normalization strategies for the MaChAmp tagger. Counts are all relative
compared to the baseline (LAI). The tag on the left is gold, right is
predicted.}
\label{tab:confMachamp}
\end{table}

\begin{table}[h!]
\centering
 \resizebox{\linewidth}{!}{
\begin{tabular}{l |r r r r}
\toprule
 & LAI & Multiling. & Lang-aware & Gold \\
\midrule
PROPN-VERB & 507 & +67 & -417 & +65 \\
PROPN-NOUN & 310 & -95 & +138 & -205 \\
VERB-NOUN & 239 & +32 & -44 & -134 \\
NOUN-ADJ & 200 & -59 & -44 & -65 \\
SYM-PUNCT & 194 & +19 & +83 & +152 \\
NOUN-PROPN & 162 & -19 & +17 & +14 \\
INTJ-NOUN & 152 & +12 & -10 & -30 \\
SYM-ADJ & 127 & -90 & -126 & -96 \\
ADJ-NOUN & 124 & +2 & -16 & -20 \\
NOUN-VERB & 122 & -12 & -27 & -39 \\
\bottomrule
\end{tabular}}
\caption{10 most common POS tagging errors for LAI baseline, counted for all
normalization strategies for the Bilty tagger. Counts are all relative compared
to the baseline (LAI). The tag on the left is gold, right is predicted.}
\label{tab:confBilty}
\end{table}

\end{document}